\documentclass[letterpaper, 10 pt, conference]{ieeeconf}  

\IEEEoverridecommandlockouts                              

\usepackage{graphics} 
\usepackage{epsfig} 
\usepackage{times} 
\usepackage{amsmath} 
\usepackage{amssymb}  
\usepackage{gensymb}
\usepackage{multirow}
\usepackage{hhline}
\usepackage{algorithm}  
\usepackage{algpseudocode}  
\usepackage{cite}
\usepackage{subfigure}       
\usepackage{graphicx}
\usepackage{xcolor}
\usepackage{url}
\usepackage{multirow}
\usepackage{bbding} 
\let\labelindent\relax
\usepackage{enumitem}
\usepackage{booktabs}
\usepackage{array, caption, threeparttable}

\setitemize{fullwidth,itemindent=\parindent,listparindent=\parindent,itemsep=0ex,partopsep=0pt,parsep=0ex}
\setenumerate{fullwidth,itemindent=\parindent,listparindent=\parindent,itemsep=0ex,partopsep=0pt,parsep=0ex}
\usepackage{dblfloatfix}

\title{\LARGE \bf
Meta Reinforcement Learning Based Sensor Scanning in 3D Uncertain Environments for Heterogeneous Multi-Robot Systems
}

\author{Junfeng Chen$^{1}$, Yuan Gao$^{1}$, Junjie Hu$^{1}$, Fuqin Deng$^{1}$ and Tin Lun Lam$^{1,2,\dagger} $
\thanks{This paper is supported by the National Key R\&D Program of China (2020YFB1313300) and Shenzhen Institute of Artificial Intelligence and Robotics for Society (AC01202101026).}
\thanks{$^{1}$Authors are with the Shenzhen Institute of Artificial Intelligence and Robotics for Society}%
\thanks{$^{2}$Authors are with School of Science and Engineering, The Chinese University of Hong Kong, Shenzhen}%
\thanks{$^{\dagger}$Corresponding author: Tin Lun Lam
        {\tt\small tllam@cuhk.edu.cn}
        }
}
\begin{document}

\maketitle
\thispagestyle{empty}
\pagestyle{empty}

 
\begin{abstract}

    We study a novel problem that tackles learning based sensor scanning in 3D and uncertain environments with heterogeneous multi-robot systems. Our motivation is two-fold: first, 3D environments are complex, the use of heterogeneous multi-robot systems intuitively can facilitate sensor scanning by fully taking advantage of sensors with different capabilities. Second, 
    in uncertain environments (e.g. rescue), time is of great significance. Since the learning process normally takes time to train and adapt to a new environment, we need to find an effective way to explore and adapt quickly.
    To this end, in this paper, we present a meta-learning approach to improve the exploration and adaptation capabilities. The experimental results demonstrate our method can outperform other methods by approximately 15\%-27\% on success rate and 70\%-75\% on adaptation speed. 


\end{abstract}

\section{INTRODUCTION}

The area of multi-robot sensor scanning studies the problem of deploying
a team of robots with sensor capabilities to maximize the observation of interest in the  environment\cite{corah2021volumetric}.
It plays a critical role in many applications, such as the 
surveillance in the disaster \cite{luo2018adaptive}, automatic patrol at sea under security requirements \cite{guo2021robust}, and thus has gained great attention over the past years. 
However, most previous works only either study sensor scanning problems in 2D and 3D curved-surface environments or use robots with similar dynamic models \cite{gupta2017cooperative,koutras2021marsexplorer,hu2020voronoi,thabit2018multi,yu2016optimal}.
Unfortunately, in 3D space, homogeneous robots with the same sensing abilities or kinetic models tend to fail due to the complexity and uncertainty of environments. 
For instance, robots on the ground are hard to scan objects that appear in 3D space.

In 3D environments, it is obvious that heterogeneous multi-robot systems have more advantages than homogeneous multi-robot systems. First, the efficiency of sensor scanning can be potentially improved by fully taking advantage of the scanning capabilities of heterogeneous robots. Similarly, the effectiveness can be also improved as long as
we can learn a good planning strategy for each robot.
Therefore, we attempt to develop a novel sensor scanning system based on heterogeneous multi-robot systems. 



\begin{figure}[t]
    \centering
    \includegraphics[width=0.95\linewidth,height=3in]{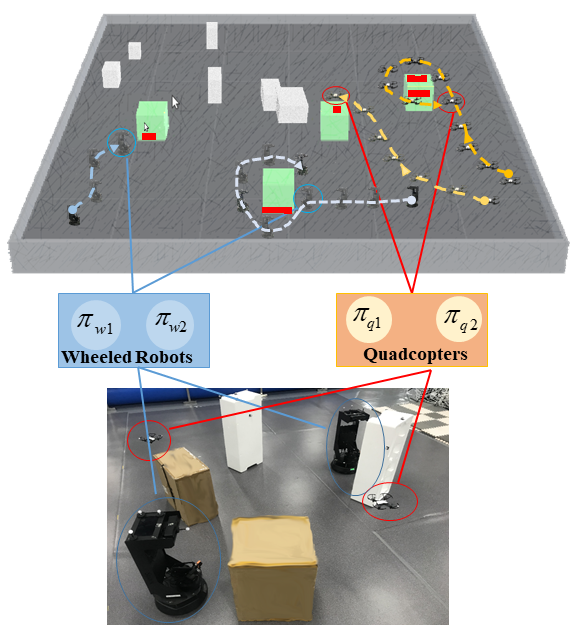}
    \centering
    \vspace{-2mm}
    \caption{An example of sensor scanning in a 3D environment for heterogeneous multi-robot systems. 
    Our method is presented to quickly adapt to uncertain environments.}
    \label{fig:first}
    \vspace{-5mm}
\end{figure}



However, many existing approaches require prior knowledge of 3D maps about the exploration environment \cite{almadhoun2019survey}. To deal with sensor scanning in unknown environments,
it is more applicable to take a reinforcement learning strategy that
firstly trains a model with numerous training episodes and then deploys it into trained environments. In practical environments, we argue that the most difficult challenge here is how to tackle the uncertainty of unknown environments.  
Typical instances of uncertainty include objects distributed randomly and unpredictably, e.g. dynamic objects, or new environments which haven't appeared in the train set.
Unfortunately, previous methods built on reinforcement learning \cite{hu2020voronoi} neither consider the adaption from the original task nor overcome the difficulty of environmental uncertainty.

\begin{figure}
    \centering
    \subfigure[The gray environment has objects with irregular shapes while the objects in green environment are all cubes.]{
    \centering
    \includegraphics[width=0.45\linewidth]{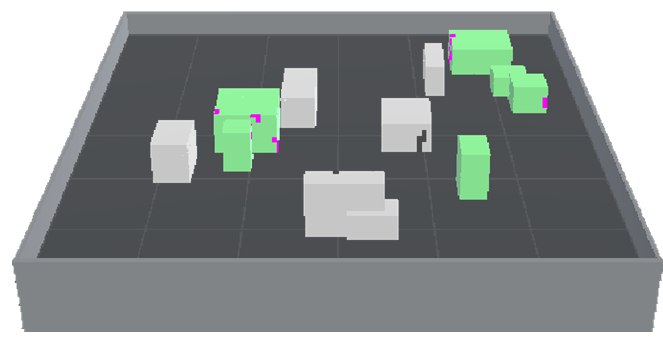}
    }
    \subfigure[The objects in the gray environment have all the same size while the objects of green environment have different sizes.]{
    \centering
    \includegraphics[width=0.45\linewidth]{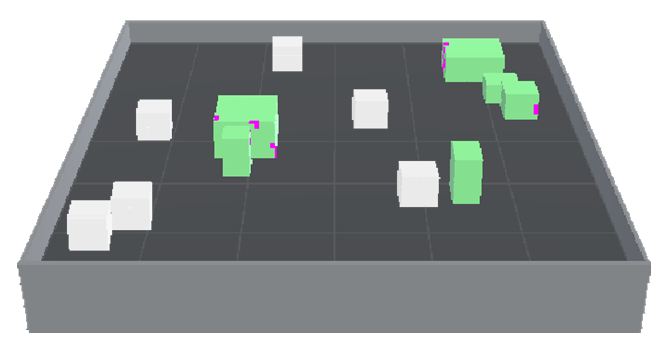}
    }
    \subfigure[There are four objects in the gray environment while there are six objects in another environment.]{
    \centering
    \includegraphics[width=0.45\linewidth]{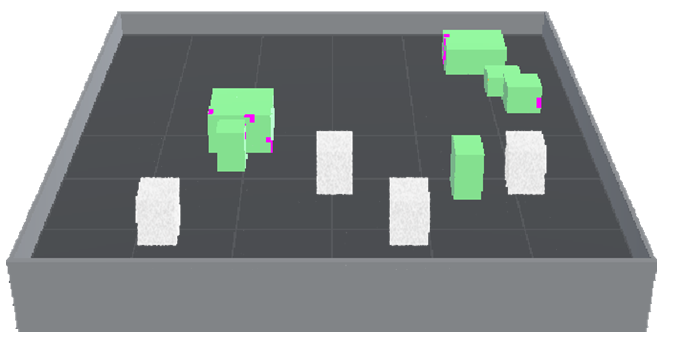}
    }
    \subfigure[The relative positions of objects in gray environment are different from those of green environments.]{
    \centering
    \includegraphics[width=0.45\linewidth]{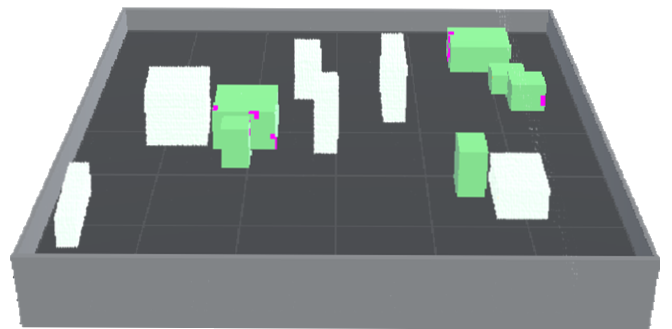}
    }
    \centering
        \caption{Examples of different environments. They are respectively comprised of green and gray objects. 
        We specifically consider four types of uncertainty, including object shape, size, number and relative positions.}
    \label{fig:second experiment}
    \vspace{-5mm}
\end{figure}

In this paper, 
we present \textit{meta-MRSS}, a meta multi-agent reinforcement learning method which is robust to the uncertainty and can be also quickly adapted to new environments. To be specific, we firstly sample the training data to form multiple tasks through the combination of different types of uncertainty. Then we apply meta multi-agent reinforcement learning methods to train an initial policy network for each robot. After that, the well-adapted initial policy network can be further trained in uncertain environments. Finally, we can achieve fast adaptation to uncertain environments through few training.
Our method is tested on numerous types of uncertainty, including number, size, shape and relative positions of objects, as shown in Fig.~\ref{fig:second experiment}.
As a result, it outperforms existing methods in both exploration efficiency and coverage effectiveness.

Our contributions are summarized as follows:
\begin{itemize}
    \item We study multi-robot sensor scanning in fully 3D environments with the proposal of using heterogeneous multi-robot systems. They perform quickly and effectively in 3D space compared to homogeneous multiple robots. 
    
    \item We present a meta multi-agent reinforcement learning based method to handle the challenge of uncertainty in unknown environments. The trained policy from our method can quickly adapt to uncertain environments and realize sensor scanning effectively.
    
    \item Experimental results show that our method outperforms other approaches by approximately 15\%-27\% on coverage success rate and about 
    70\%-75\% on adaptation speed.

\end{itemize}

\section{Related Work}
\subsection{Multi-Robot Sensor Scanning}

In the related literature, many works focus on the problems of coverage control in 2D space and 3D-curved surface environments rather than fully 3D space \cite{GALCERAN20131258}. 
Towards the 2D and 3D curved surface environments, \cite{jamshidpey2021centralization} introduces a general control structure for the homogeneous multi-robot system in the 2D environment which is hard to be directly utilized in 3D space.  In \cite{7487622, 8588984,wang2012energy,melo2021dynamic, santos2018coverage}, all methods are based on Voronoi graph. However, Voronoi based methods do not effectively divide the 3D space into sub-regions or are not adaptive to dynamic and uncertain environments.
In \cite{cheng2008time}, the authors present an online time-optimal method to scan 3D urban structures. This method considers a completely structured and static environment, however, it can not efficiently cope with the environmental uncertainty. In \cite{englot2012sampling}, the authors introduce an offline sampling method to inspect the complex surface of the ship hull, while the presented algorithm is not able to handle the difficulties of 3D space.

Regarding 3D space, the most similar work to ours is \cite{8205999}, however, this method only considers single robot exploration in the fixed environment which will lead to low efficiency compared with multi-robot systems. In \cite{cheng2008time,bircher2016three}, the authors propose a re-sampling and time-optimal approach to plan trajectory for 3D urban structure \cite{cheng2008time}, while they only consider the single aerial robot which has a clear disadvantage compared with multiple robots.



\subsection{Sensor Scanning under the Uncertainty}
In the application of sensor scanning, we must carefully consider the uncertainty, as it has a huge effect on scanning efforts. Recently, most works focus on the uncertainty of sensors and localization. In \cite{7407311}, the authors develop a coverage-optimization method under the localization and sensing uncertainty. However, this uncertainty occurs in one specific scenario without considering environmental uncertainty. In \cite{6094695}, the authors propose a probability method to predict the uncertainty in sensing and actuation in the specific free space. Similarly, an adaptive collaboration method for multi-robots under localization uncertainty is presented in \cite{8816356}. In \cite{8431115}, the authors address the problem of the inaccurate position measurement in deploying the heterogeneous multi-robot system to cover a given environment. Unfortunately, all of these methods consider one fixed environment, when faced with environmental uncertainty, tend to fail.

Another kind of uncertainty that is concerned by researchers is dynamic objects because their motion model is hard to predict. Therefore, many works focus on how to deal with uncertainty in the face of unpredictable obstacles while performing tasks.
Some prior works
\cite{dai2020graph, shokry2020leveraging} mainly utilize reinforcement learning to learn an end-to-end coverage planning policy to plan a complete coverage path while avoiding the dynamic obstacles. 
When facing a previously unseen task, these trained models in the uncertain environments will yield bad performance due to poor generality and will be extremely time-consuming to learn from scratch.

\begin{figure*}[hbt!]
    \centering
    \includegraphics[width=0.99\linewidth]{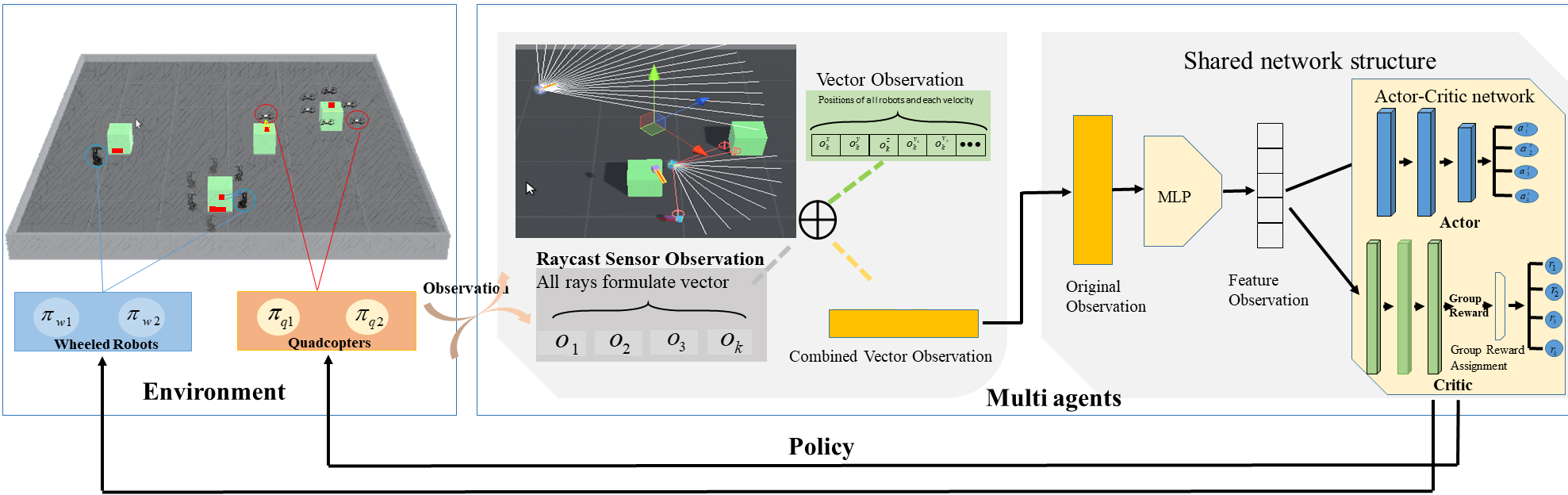}
    \centering
    \vspace{-2mm}
    \caption{The structure of meta optimizer. The wheeled robots and quadcopters use different sets of policy networks. 
    They utilize decentralized actor and centralized critic networks.
    This method is well applicable to heterogeneous multi-robot systems.}
    \label{fig:framwork}
    \vspace{-5mm}
\end{figure*}

\subsection{Meta Reinforcement Learning}
To deal with the diversity and uncertainty of the environments, how to fast adapt into new environments has aroused huge attention by researchers. It is generally accepted that meta-Learning \cite{lemke2015metalearning} can be fast adapted into new environments while maintaining good performance in uncertain environments.
  The Meta-Learning method called MAML is proposed to quickly solve multi-tasks of cheetah locomotion in \cite{finn2017model}. Inspired by this work, our approach is proposed to solve the problem of fast adaptation of multi-robot systems in the presence of uncertainty. In \cite{li2020uav}, Meta-TD3 is proposed to help UAVs to quickly adapt to new target movement mode and obtain better tracking effectiveness. In \cite{ghadirzadeh2021bayesian}, the authors propose a Bayesian meta-learning method to quickly adapt to different robotics platforms.

The key difference is that
previous methods of meta reinforcement learning are built on a single robot, while our method is built on heterogeneous multi-robot systems. We aim to learn a good strategy to enable effective collaboration and fast adaptation to uncertain environments.

\section{Methodology}
In this section, we first model environmental uncertainty. Then we advance our method to provide good initial policy networks for uncertain environments.

\subsection{Problem Formulation}
We specify the scientific problem of heterogeneous multi-robot sensor scanning as whether objects are scanned by these robots within different sensing abilities in 3D and uncertain environments. 

As discussed in \cite{adepegba2016multi,8205999}, we consider a group of $N_R = \{R_w \cup R_q\}$ heterogeneous robots, in which wheeled robots represented by $R_w$ carrying a cube-view sensor with a sensing range of 2 meters can only move on the ground, 
while quadcopters labeled by $R_q$ with a 3.5-meter-range sensor could fly in 3D space to scan the whole discretized environment $\Omega$ comprised of 3D objects $\Phi_i$, namely $\Omega = \{\Phi_1, \Phi_2, \cdots, \Phi_n \}$ with uncertainty $G$. We assume that the uncertainty $G$ refers to the relative positions, number, shape and size of objects $\Phi_i$. As for each $\Phi_i$, we assume it consists of cells $\mathcal{C} = \{1,2,\cdots,m\}$ whose state is either unoccupied (\textbf{0}) or occupied (\textbf{1}) ($ c_i \in \{0,1\}$), according to whether the cell is scanned. We assume that in these factors, relative positions and shape satisfy normal distribution while size and number obey randint distribution. Robots $N_R$ move around the environments with states $ S_{t,R} \in \mathbb{R}^3 $ at discrete time intervals via the dynamic models
\begin{equation}
    S_{t,R} = f(S_{t-1,R}, U_{t,R})
\end{equation}
Specifically, the state of the wheeled robots $R_w$ is $S_{t,R_w} = [x_{t,R_w}, y_{t,R_w}, 0]$. However, the state of the quadcopters $R_q$ exists in 3D environments $S_{t,R_q} = [x_{t,R_q}, y_{t,R_q},z_{t,R_q}] $. Additionally, where $U_{t,R} \in \mathcal{U}$ belongs to a finite set of control inputs, referring to velocity command in our settings. Because of safety requirements, multi-robot systems must remain a safety scanning region to avoid colliding with other robots or objects. Therefore, we should constrain the states of robots into a safety scale
\begin{equation}
    S_{t,R} \in \mathcal{X}_{safe}(\Omega)
\end{equation}
Assuming that scanning measurement, including scanning range and viewpoint, is deterministic, without adding Gaussian noise, robots could determine definitely whether the cells of objects are scanned within effective scanning range. Therefore, when scanning the cells within above sensor ranges, robots can infer the number of occupied cells $\Theta^{sensor}(S, \Omega) \subseteq \mathcal{C}$ which are rewarded as occupancy values
\begin{equation}\label{eqa:sensor}
    y_{t,R} = h(S_{t,R},\Phi) = \{(i, c_i): i\in \Theta^{sensor}(S, \Omega) \}
\end{equation}
In Eq.~\ref{eqa:sensor}, reward $y_{t,R}$ is proportional to $\Theta^{sensor}(S, \Omega)$. The objective is that heterogeneous multi-robot systems try to scan as many cells of objects in uncertain environments as possible in a given time horizon:

\begin{figure}[t]
    \centering
    \includegraphics[width=0.95\linewidth,height=1.2in]{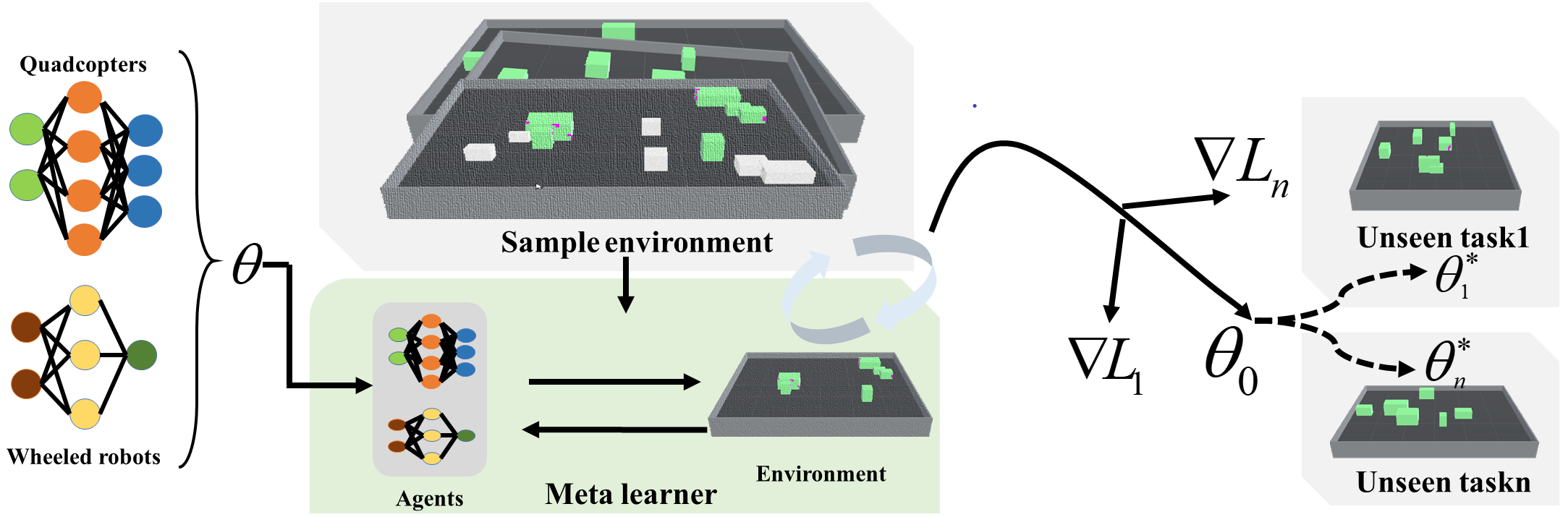}
    \centering
    \vspace{-2mm}
    \caption{The structure of meta-MRSS. Firstly, meta policies are obtained through meta-MRSS. Then the optimal policies for uncertain environments could be learned through few training of meta policies.}
    \label{fig:maml}
    \vspace{-4mm}
\end{figure}

\begin{equation}\label{equation}
\begin{aligned}
   & \underset{\Omega \in G}{\max} \sum_{t' \in \{1,2,\cdots, t\}, R \in N_R} \Theta^{sensor}(S_{t\prime, R},\Omega) \\
   & \mbox{s.t.}\quad 
    S_{t', R} \in \mathcal{X}_{safe}(\Omega_{t'}) \\
    & \qquad S_{t', R} = f(S_{t', R}, U_{t', R})\\
    & \qquad y_{t', R} = h(S_{t', R}, \Theta)\\
    & \qquad for \ all \ t' \in \{1,2, \cdots, t\} and \ R \in N_R
    \end{aligned}
\end{equation}


\subsection{meta-MRSS}
In this section, 
our method as shown in Algorithm.~\ref{alg::maml-poca} is divided into two parts, respectively meta optimizer in the inner loop and meta learner in the outer loop. Firstly, in the inner loop, we make sure that meta optimizer could converge in specific environments. Then we sample various train environments into multiple tasks, by meta learner we could obtain good-performance meta policy which can be used to learn a model in uncertain environments with few training to realize fast adaptation. 
\renewcommand{\algorithmicrequire}{\textbf{Input:}}  
\renewcommand{\algorithmicensure}{\textbf{Output:}} 
\begin{algorithm}[t]
  \caption{meta-MRSS} 
  \label{alg::maml-poca}
  \begin{algorithmic}[1]
    \Require
      p($\tau$): samples from uncertainty distribution
    \Require
      $\mathbf{I} $: meta-update iterations 
    \Ensure
      $\pi_{0}$: meta policy
    \State randomly initialize $\theta_j \in \{\theta_1 \cup \theta_2 \}$
    \For{all iteration in $\mathbf{I}$}
      \State Sample buffers of tasks $\tau \sim p(\tau)$
      \For{$\tau_i$ in $\tau$}
        \State Sample $\mathbf{K}$ trajectories $\mathbb{D}$ using $f_{\theta_j}$ in $\tau_i$
        \State Calculate loss $\mathcal{L}_{\tau_i}$ using $\mathbb{D}$ via meta optimizer in

        \quad Eq.(~\ref{eq:loss}).
        \State Compute adapted parameters by optimizer: $\theta'_{j,i}$
        \State Sample validation trajectories respectively $\mathbb{D}'_{j,i}$ 
        
        \quad using $f_{{\theta}'_{j,i}}$ in $\tau_i$
      \EndFor
      \State Update $\theta_j$ via $\sum_{\tau_i \sim p(\tau)}\mathcal{L}_{\tau_i}(f_{\theta'_j})$ using $\mathcal{D}'_i$ and $\mathcal{L}_{\tau_i}$ 
    \EndFor
  \end{algorithmic}
 \end{algorithm}

\subsubsection{Meta optimizer}
To achieve this problem modeling, we build the model in the simulation through the toolkit called \textit{mlagents}\cite{juliani2020unity}. As shown in Fig.~\ref{fig:framwork}, we build some green objects that will be scanned by sensors while green tile-like objects are tightly attached in five planes of objects to simulate cells. 

It is generally accepted that MDP for multi-agent systems could be defined as: $\mathcal{M} = <\mathcal{A}, \mathcal{O}, \{\mathcal{U}_i\}_{i \in \mathcal{A}},f, \{p_i\}_{i \in \mathcal{A}} > $, where $\mathcal{A} = \{1,2,\cdots, n\}$ is a set of agents and $\mathcal{O}$ represents the global state of the whole environment. Then a set of joint actions are selected from action space $\mathcal{U} = \times_{i \in \mathcal{A}}\mathcal{U}_i$. We seek the optimal joint policies $\mathbb{\pi}^{*} = \{\mathcal{A}\times \mathcal{U}\} \rightarrow{\mathbf{R}}$, where $\mathbf{R}$ is the function of reward. As for multi-agent systems, communication should be most concerned. We build the models in five aspects: 
\begin{itemize}
    \item \textit{Joint Action space}: 
    To make robots' motion suitable for real-world applications, we set joint continuous action space. In specific, we set forward-back, the left-right velocity of wheeled robots with  $-0.2\sim0.2$ m/s and turn velocity of $-3\sim3$ rad/s. The main difference between quadcopters and wheeled robots is that forward-back, left-right, up-down velocity is $-0.1\sim0.1$ m/s. Besides, we also set the pitch and yaw velocity of quadcopters both are $-3\sim3$ rad/s. 
    
    \item \textit{Global observation space}: In the simulation, we deploy raycast sensors to collect observations. These raycast sensors emit rays every 5 degrees within a limited scanning range. When hitting the objects, hit rays will return the positions and state of cells. As shown in Fig.~\ref{fig:framwork}, to facilitate the learning process, we adopt one-hot ways to encode the state of the cells into 0 (unoccupied) or 1 (occupied) in our setting. 
    Besides the observation of sensors, we also collect the positions and velocity as observation input. A detail about the observation as positions and velocity will be further discussed in the below part. 
    
    \item \textit{Communication setting}: 
    To achieve good-performance collaboration for multi-robot systems, we also consider the communication mechanism. We conduct some ablation experiments (see the ablation study in the experiment) to demonstrate a good mechanism in which wheeled robots could access each position of all robots and their own velocity while quadcopters only send the object information to wheeled robots.
    \item \textit{Reward Shaping}: To achieve collaboration, the reward is defined as follows:
    \begin{equation}
    \mathbf{R} = 
    \begin{cases}
    1& \text{if robots scan the cell}\\
    -1& \text{if robots collide with objects}\\
    -1& \text{if robots collide with other robots}\\
    \end{cases}
\end{equation}
where we assume when wheeled robots or quadcopters scan one cell they will be rewarded one point. To avoid obstacles including objects and other robots, if they collide, they will be punished one point. 

\item \textit{Training method}: 
As shown in Fig.~\ref{fig:framwork}, we configure two sets of policy networks respectively for wheeled robots and quadcopters. Each network follows the actor-critic structure with central critics and distributed actors. We combine the observation vector of raycast sensors and another observation vector of robots' position and velocity as a combined observation vector. To reduce observation dimension and improve training efficiency, 3-layer MLP is adopted to encode combined observation vector to output feature observation space, finally input it into the actor and critic network. Inspired by the POCA algorithm \cite{lowe2020multiagent, foerster2018counterfactual}, 
 the team members share one reward which is assigned into different reward values according to the contribution members make in this group. 
     
\end{itemize}

\subsubsection{Meta learner}

As shown in Fig.~\ref{fig:maml}, meta-MRSS utilizes MAML structure as meta learner, then we calculate the loss advanced by Eq.~\ref{eq:loss} to update network parameters.

\begin{equation}\label{eq:loss}
\begin{aligned}
&\mathbb{A}^a(s,u) = Q(s,u) - \sum{\substack{u'^a}}\pi^a(u'^a|\tau^a)Q(s,(u^{-a},u'^a))\\
& \qquad \ \mathcal{L}_{\tau_i}(f_{\theta_j,i}) = - 
    \mathbf{E}_{x_t, a_t \sim f_{\theta_j,i}}[\mathbb{A}^a(s,u)]\\
\end{aligned}
\end{equation}
where $\mathbb{A}^a(s,u)$ represents the counterfactual baseline of the single agent labeled by $a$, $u^a$ is each agent's action, $\tau^a$ refers to the past action sequence. While $u$ is the joint actions of all agents, $u^{-a}$ presents the joint actions of other agents, $s$ represents the global state.

In the inner loop, we use more than one gradient descent update due to the dynamic and complexity of multi-agent systems and update network parameters using Adam optimizer. The illustration is shown in Algorithm~\ref{alg::maml-poca}.

\section{Experiments}
We firstly provide details of the experimental settings including train set, test set and hyperparameters. We then demonstrate the performance of our proposed method in the test set by comparing with three baselines methods. Since the effect of meta optimizer has a direct impact on the performance of our approach, we finally discuss the factors that affect the meta optimizer, including reinforcement learning algorithm selection, communication mechanisms, and multi-robot system homogeneity or heterogeneity. 



\subsection{Quantitative Evaluation}
 \subsubsection{Experimental Setup}
 
 As shown in Fig.~\ref{fig:second experiment}, to fully simulate the uncertain environments, we randomly sample 1000 environments as a train set, then we sample 5 unknown environments as the test set to validate that our method could outperform in the test set through fewer training. To make a fair comparison, 
 we train these networks under the same hyperparameters. In detail, we set a part of the important hyperparameters of meta-MRSS presented in Algorithm.~\ref{alg::maml-poca} as follows:
 
\begin{table}[hbt!]
\centering
\begin{tabular}{ccccccccc}
\hline
hyper  parameters &$\mathbf{I}$    &$\tau$  &$\mathbf{K}$   &$\beta$  &$\epsilon$   &$\lambda$  &epoch\\ \hline
value           & 100 & 10  & 20  &0.005 &0.2 &0.95 &3 \\ \hline
\end{tabular}
\captionsetup[table]{singlelinecheck=false}
\caption{Hyperparameter setting. $\mathbf{I}$, $\tau$ and $\mathbf{K}$ are explained in the meta-MRSS. $\beta$ represents the strength of the entropy regularization. $\epsilon$ influences how rapidly the policy evolves. $\lambda$ depicts the regularization parameter.}
\vspace{-4mm}
\end{table}

\subsubsection{Baselines Methods} 
 To quantify our method, 
we provide a thorough evaluation of our method by comparing it with existing methods. 
To be specific, we choose the following three methods dealing with the uncertainty as baselines:
 \begin{itemize}
     \item \textbf{Standard POCA} Since our method is built on POCA. Therefore, we conduct experiments to fairly compare our method against the standard POCA. We utilize standard POCA in the test set to obtain the average rewards.
     
     \item \textbf{Domain Randomization}
     Domain Randomization technique
     \cite{tobin2017domain} is a common way to improve the generality of reinforcement learning. Therefore, we simulate environmental uncertainty through simulators, then use the POCA algorithm in these random simulated environments to obtain a model with high generalization capability.
     
     \item \textbf{Transfer Learning}
     Considering that most of the tasks are correlated, transfer learning  \cite{torrey2010transfer} allows to share the learned model parameters for the new models to accelerate and optimize the learning efficiency. We first obtain the initial networks on the train set and then perform a second training on the test set to obtain the average rewards. 
     
 \end{itemize}

 \subsubsection{Experimental Results}
\begin{figure}[hbt!]
    \centering
    \subfigure[The learning process of the wheeled robots.]{
    \centering
    \includegraphics[width=0.95\linewidth]{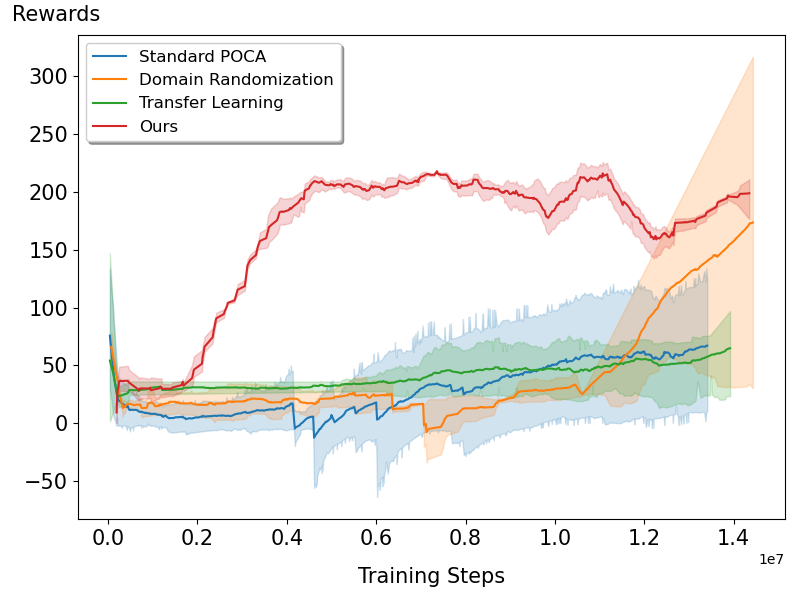}
    }
    \subfigure[The learning process of the quadcopters.]{
    \centering
    \includegraphics[width=0.95\linewidth]{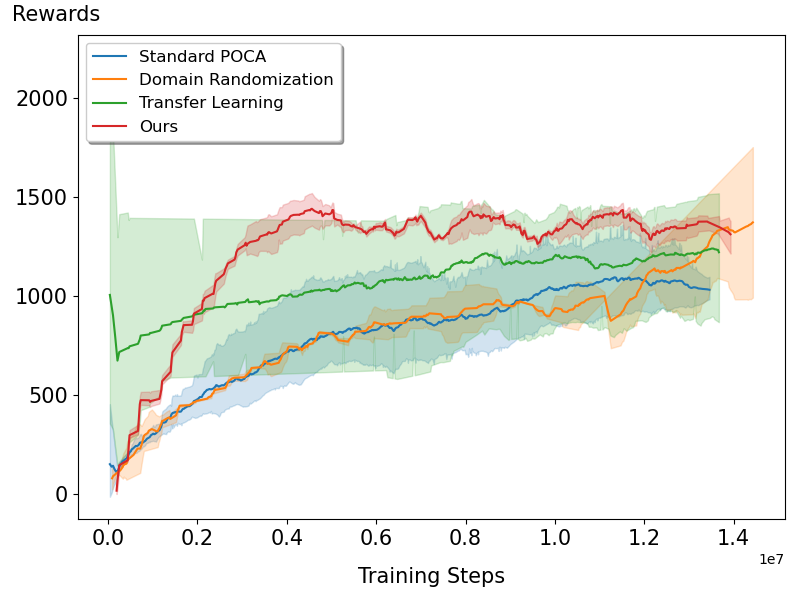}
    }
    \subfigure[The scanning success rate of four methods.]{
    \centering
    \includegraphics[width=0.95\linewidth]{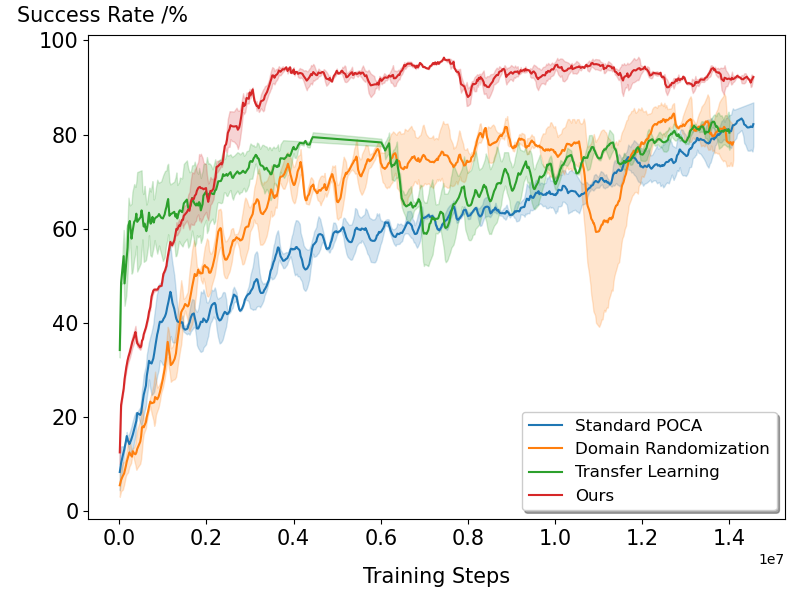}
}
    \centering
    \caption{The comparison results of learning for two kinds of robots using respectively four methods. The results present meta-MRSS has better learning efficiency ans success rate in the face of uncertain environments.}
    \label{fig:final experiment}
    \vspace{-7mm}
\end{figure}

 To compare the performance of these baselines methods and meta-MRSS, 
 we use the average reward and scanning success rate in a test set through three trials as metrics. The evaluation results are respectively shown in
 Fig.~\ref{fig:final experiment}.
As indicated by the results in Fig.~\ref{fig:final experiment} (a) \& (b), our method can converge in almost 4 million steps, while other methods require 14 million steps to converge. 
Statistically, our method can outperform other baselines methods approximately 70\%-75\% on the convergence speed.
It proves that our method can quickly learn a new model to adapt to uncertain environments.

In the scanning success rate, Fig.~\ref{fig:final experiment} (c) turns out that our method can achieve a better scanning performance in the test set. Statistically, our method can reach an average success rate approximately of 87.12\%, Domain Randomization based POCA can reach 68.54\%, Standard POCA's success rate is 60.16\% and Transfer Learning can reach 71.74\%. We observe that although Transfer Learning has a higher scanning success rate than other methods at the beginning, when facing uncertain environments, it needs time to fine-tune the networks which can lead to the steep drop in scanning efforts. It is obvious that the curves labeled by Standard POCA slowly increase because POCA has weak generality in uncertain environments. Domain randomization-based method efficiently improves the generality of POCA, however, it has a lower level of scanning success rate by comparing with our method in given steps.
\subsection{Ablation Study}
\vspace{0mm}
In this section, we conduct three comparison experiments. First, we demonstrate POCA can be well suitable as the meta optimizer in our method. Second, we verify the effect of communication mechanisms on the convergence performance of POCA. Third, we prove the merit of heterogeneous multi-robot systems for sensor scanning tasks in 3D environments by comparing with homogeneous multi-robot systems.
We set up one environment to do ablation studies, specifically, four green objects, which have a size of $6 \times 6 \times 6$ and are attached into 180 cells, are randomly placed in the specific region. The goal of this task we define is how heterogeneous multi-robot systems scan all the objects as quickly as possible while avoiding the obstacles. 

\begin{itemize}
    \item \textbf{Communication Mechanism}: 
    We test seven schemes in the above specific environment to see if reward curves reach convergence. The experimental results demonstrate that the first scheme requires 8 million steps to converge while the sixth scheme takes 14 million steps to reach convergence. Qualitatively, Tab.~\ref{tab::ablation} also demonstrates that the first scheme we choose yields better performance. It makes sense that wheeled robots have comparatively limited-range sensors, so they require the observation information of other wheeled robots and quadcopters to scan undetectable objects. 
    
    \item \textbf{POCA vs. PPO}: 
    To make a fair evaluation, we conduct the comparison experiment using the same hyperparameters with PPO which is the useful algorithm in the robotics field. 
    The comparison result of PPO and POCA is presented in Fig.~\ref{fig:first-result}. 
   Training curves by POCA have good convergence performance. 
    Therefore, POCA could be implemented as a meta optimizer in our method.
    


    \item \textbf{Heterogeneous vs. Homogeneous multi-robot systems}:
    We utilize two wheeled robots and two quadcopters as heterogeneous multi-robot system while we use four quadcopters as homogeneous multi-robot system. 
    In addition, we run three experiments in the same settings to calculate the average reward for two kinds of multi-robot systems as the evaluation metrics.

 The comparison results are shown in Fig.~\ref{fig:homovshetero}. It turns out that heterogeneous multi-robot systems can scan as many objects as possible in a given steps. It further demonstrates that heterogeneous multi-robot systems are more advantageous than homogeneous robots in 3D environments.
 
 \begin{table}[hbt!]
\setlength{\tabcolsep}{2mm}{
\begin{tabular}{cllllcl}
\hline
\multicolumn{1}{l}{\multirow{2}{*}{Scheme}} & \multicolumn{2}{c}{Wheeled robots} & \multicolumn{2}{c}{Quadcopters} & \multicolumn{2}{c}{\multirow{2}{*}{\begin{tabular}[c]{@{}c@{}}Convergence\\  results\end{tabular}}} \\ \cline{2-5} 
\multicolumn{1}{l}{} & positions & velocity & positions & velocity &\multicolumn{2}{c}{} \\ \hline
1   & $\checkmark$               & $\checkmark$                & $\times$                & $\times$             & \multicolumn{2}{c}{\textbf{succeed}}                                                               \\ \hline
2   &$\times$                  &$\checkmark$                 & $\times$                & $\times$              & \multicolumn{2}{c}{failed}                                                                         \\ \hline
3   &$\times$                  & $\times$                &$\times$                 &$\times$               & \multicolumn{2}{c}{failed}                                                                         \\ \hline
4   & $\checkmark$                 & $\times$                &$\times$                 &$\times$               & \multicolumn{2}{c}{failed}                                                                         \\ \hline
5   & $\checkmark$                 &$\checkmark$                 &$\checkmark$                 &$\times$               & \multicolumn{2}{c}{failed}                                                                         \\ \hline
6   & $\checkmark$                 &$\times$                 & $\checkmark$                &$\checkmark$               & \multicolumn{2}{c}{\textbf{succeed}}                                                                \\ \hline
7   &$\checkmark$                  &$\checkmark$                 &$\checkmark$                 &$\checkmark$               & \multicolumn{2}{c}{failed} \\ \hline                                                             
\end{tabular}}
\captionsetup[table]{singlelinecheck=false}
\caption{The results of communication schemes. $\checkmark$ and $\times$ refers to whether robots communicate related information. 'failed' means that the learning curves hardly converge. 'succeed' represents the learning curves can converge.}

\setlength{\tabcolsep}{7mm}
\vspace{-2mm}
\label{tab::ablation}
\end{table}

\begin{figure}[hbt!]
    \centering
    \includegraphics[width=0.9\linewidth]{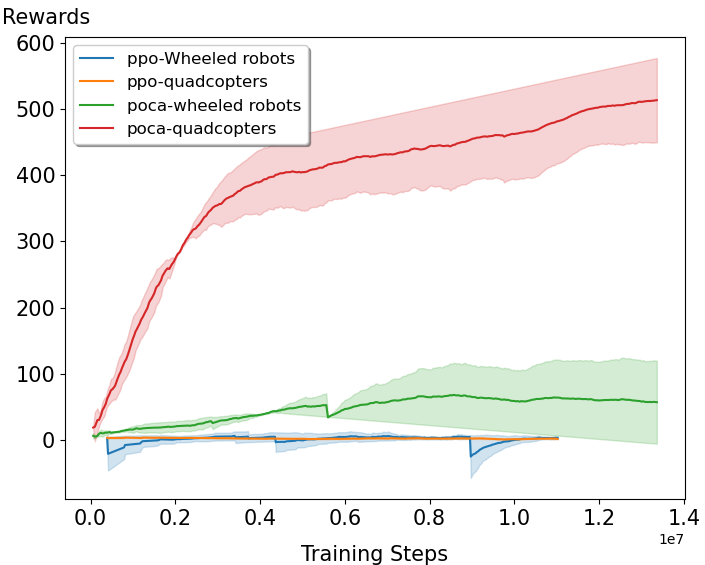}
     \caption{The comparison results. The solid curves and shaded regions depict the mean and standard deviation of average reward among three trials. We can observe the POCA could converge well for wheeled robots and quadcopters.
    }
    \label{fig:first-result}
    \vspace{-3mm}
\end{figure}

 \begin{figure}[hbt!]
    \centering
    \includegraphics[width=0.9\linewidth]{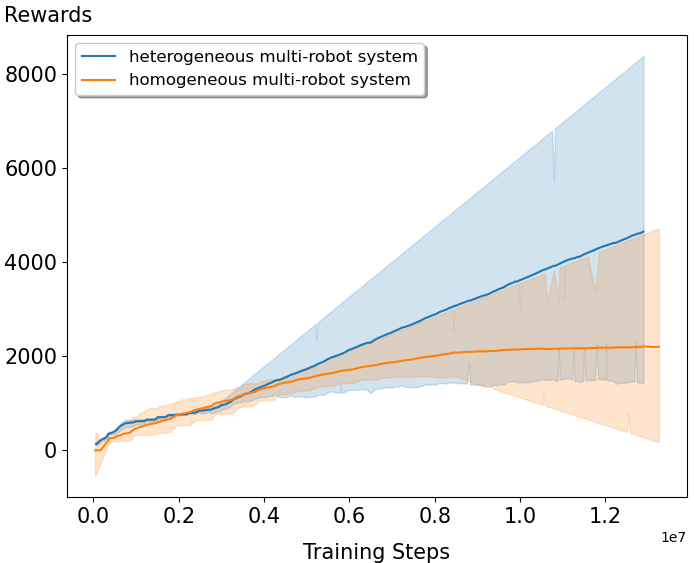}
     \caption{The comparison results. It turns out that heterogeneous multi-robot systems can obtain more rewards than homogeneous robots in the same steps. 
    }
    \label{fig:homovshetero}
    \vspace{-6mm}
\end{figure}

\end{itemize}

\section{CONCLUSIONS}
In this paper, we study the problem of how heterogeneous multi-robot systems could collaborate to perform sensor scanning in 3D uncertain environments. We attempt to handle the uncertainty that appears in unknown environments with a proposed method based on meta multi-agent reinforcement learning.  
Experimental results demonstrate that our method outperforms other approaches in efficiency and effectiveness of sensor scanning tasks.


\bibliographystyle{IEEEtran}
\bibliography{reference.bib}
\end{document}